\def\year{2022}\relax
\documentclass[letterpaper]{article} 
\usepackage{aaai22}  
\usepackage{times}  
\usepackage{helvet}  
\usepackage{courier}  
\usepackage[hyphens]{url}  
\usepackage{graphicx} 
\usepackage{natbib}  
\usepackage{caption} 
\DeclareCaptionStyle{ruled}{labelfont=normalfont,labelsep=colon,strut=off} 
\usepackage{algorithm}
\usepackage{algorithmic}
\usepackage{framed}

\usepackage{latexsym}
\usepackage{amsmath}
\usepackage{amsfonts}
\usepackage{graphicx}

\usepackage{subfigure}
\usepackage{colortbl}
\usepackage{arydshln}
\usepackage{xcolor}
\usepackage{multirow}
\usepackage{bm}
\usepackage{makecell}
\usepackage{newfloat}
\usepackage{listings}
\floatstyle{ruled}
\newfloat{listing}{tb}{lst}{}
\floatname{listing}{Listing}
\title{Topic Driven Adaptive Network for Cross-Domain Sentiment Classification}
\author{
    Yicheng Zhu, \textsuperscript{\rm 1},
    Yiqiao Qiu, \textsuperscript{\rm 1},
    Qingyuan Wu, \textsuperscript{\rm 2}\thanks{Qingyuan Wu is corresponding author},
    Fu Lee Wang, \textsuperscript{\rm 3},
    Yanghui Rao, \textsuperscript{\rm 1}
}
\affiliations{
    \textsuperscript{\rm 1} Sun Yat-sen University
\\
    \textsuperscript{\rm 2} Guangdong University of Finance
\\
    \textsuperscript{\rm 3} Hong Kong Metropolitan University
\\

    zhuych9@mail2.sysu.edu.cn, qiuyq7@mail2.sysu.edu.cn, wuqingyuan@gduf.edu.cn ,
pwang@hkmu.edu.hk, raoyangh@mail.sysu.edu.cn
}

\begin{document}

\maketitle

\begin{abstract}
Cross-domain sentiment classification has been a hot spot these years, which aims to learn a reliable classifier using labeled data from the source domain and evaluate it on the target domain. In this vein, most approaches utilized domain adaptation that maps data from different domains into a common feature space. To further improve the model performance, several methods targeted to mine domain-specific information were proposed. However, most of them only utilized a limited part of domain-specific information. In this study, we first develop a method of extracting domain-specific words based on the topic information. Then, we propose a Topic Driven Adaptive Network (TDAN) for cross-domain sentiment classification. The network consists of two sub-networks: semantics attention network and domain-specific word attention network, the structures of which are based on transformers. These sub-networks take different forms of input and their outputs are fused as the feature vector. Experiments validate the effectiveness of our TDAN on sentiment classification across domains.
\end{abstract}
\section{Introduction}

Sentiment classification aims to automatically predict sentimental labels of given documents. With the development of deep learning, sentiment classification has achieved outstanding performance \cite{DBLP:journals/widm/ZhangWL18}. However, traditional sentiment classification methods based on supervised learning rely on a large amount of manually annotated data, and constructing such datasets is a laborious and expensive process. To address this problem, cross-domain sentiment classification has been proposed \cite{pan2010cross}. It eases the need for a large amount of labeled data in the target domain (where labeled data are few) by training the classifier with labeled data from the source domain (where labeled data are rich).

Recent works on cross-domain sentiment classification mainly focused on domain-invariant features extraction, also known as domain adaptation \cite{peng2018cross,li2018hierarchical,zhang2019interactive}. Data from both domains are mapped into a common feature space where all data points are assumed to be independent and identically distributed. Most works trained a classifier based on the domain-invariant features extracted from the source domain. Specifically, \citet{peng2018cross} explicitly restricted the discrepancy between features from both domains and some other works \cite{li2018hierarchical,zhang2019interactive} adopted adversarial methods to extract the domain-invariant features by confusing a domain discriminator.

To further improve the performance, methods of exploiting additional information have been proposed \cite{li2018hierarchical,peng2018cross,he2018adaptive,hu2019domain,DBLP:journals/corr/BahdanauCB14}. Among these works, \citet{peng2018cross} and \citet{he2018adaptive} introduced semi-supervised learning methods to utilize the target domain information. \citet{hu2019domain} and \citet{zhang2019interactive} leveraged the aspects of documents to assists sentiment classification. \citet{li2018hierarchical} designed a hierarchical attention network that automatically finds pivots, i.e., domain-shared sentiment words, and non-pivots, i.e., domain-specific sentiment words. Both pivots and non-pivots help in cross-domain sentiment classification.

From the previous works, we can conclude that there are two types of words in the cross-domain sentiment classification task: domain-shared words and domain-specific words. Domain-shared words occur in both domains. Most words that directly reflect sentiment polarity are shared between domains, such as \textit{good} and \textit{bad}. Domain-shared words that reflect sentiment polarity are also known as pivots \cite{li2018hierarchical}. Domain-specific words occur in a single domain and most of them don't carry sentiment information with them. Most domain-specific words describe the background information of documents, such as \textit{book} and \textit{internet}. However, domain-specific words can also include sentiment words, i.e., non-pivots \cite{li2018hierarchical}. Therefore, there are two kinds of domain-specific words: non-pivots and background words. Non-pivots are sentiment words and background words that provide us with background information about a document.

The utilization of domain-shared words has been widely explored in previous works \cite{blitzer-etal-2006-domain,blitzer7domain,li2018hierarchical,li2020simultaneous}. However, domain-specific words are not thoroughly utilized. Previous works typically only pay attention to one type of domain-specific word. HATN \cite{li2018hierarchical} only pays attention to domain-specific words that reflect sentiment polarity (i.e., non-pivots) while IATN \cite{zhang2019interactive} only utilizes the aspects of documents to provide background information without considering non-pivots. In fact, both types of domain-specific words provide useful information for sentiment classification because non-pivots provide direct sentiment information and background words would assist the semantics network to decide which aspect should be focused on, as illustrated in \citet{zhang2019interactive}.

To utilize both types of domain-specific words, we introduce topics to extract domain-specific words before training the classification network. The benefit of using topics to identify domain-specific words in two domains is that such words can be better captured compared with methods based on word co-occurrences or mutual information. Firstly, topics can help extract both types of domain-specific words while methods based on word co-occurrences or mutual information might fail to capture both types. Secondly, the proposed method helps capture polysemous words such as ``right". When ``right" means ``correct", it is a domain-shared word. On the other hand, the above word is domain-specific when it means ``interest". The topic model performs soft clustering on words which means that the same word occurring in different topics may hold different semantics and this helps to separate the different semantics of the polysemous words, deciding whether it is domain-shared or domain-specific. Particularly, we calculate topic occurrence possibilities in two domains to decide whether a topic is specific to a domain and check every word's most related topic to decide whether it is domain-specific. Domain-specific words are encoded using the domain-specific word attention network to generate feature vectors. The reason why we do not extract domain-shared words is that previous works \cite{li2017end,li2018hierarchical} showed that attention-based networks combined with domain adaptation methods would automatically capture pivots. Therefore, the semantics sub-network in our work that takes original documents as input can utilize domain-shared words, and an extra sub-network that takes domain-shared words as input is unnecessary.

Based on the extracted domain-specific words, we propose a \textit{Topic Driven Domain Adaptation Network} (TDAN) for cross-domain sentiment classification. TDAN incorporates domain-specific information to enhance the model performance. It consists of two sub-networks: semantics attention network and domain-specific word attention network. The sub-networks use different forms of inputs and the output vectors of each network are fused as the final feature vector. Then, the network adopts the adversarial method \cite{NIPS2014_5ca3e9b1} to map the feature vector into a domain-invariant space.

The main contributions are summarized as follows:

\begin{itemize}
  \item Our work is the first one to utilize the topic model to extract domain-specific words.
  \item Our work is the first one to utilize both types of domain-specific words for cross-domain sentiment classification.
  \item We develop a novel network that combines different forms of input to classify sentiments across domains.
\end{itemize}
\section{Related Work}

\textbf{Domain adaption:} The fundamental problem of cross-domain learning is that the source and target domains hold different data distributions. Directly employing a classifier trained with source domain data shows low performance because of the varied data distributions in the target domain. Therefore, domain adaptation is proposed to reduce the discrepancy between two distributions. In a preliminary study, \citet{blitzer-etal-2006-domain} proposed Structural Correspondence Learning (SCL) to classify documents utilizing human-selected pivot feature. The pivot features are shared words between both domains that indicate opinions. Some works \cite{pan2010cross,gu2011joint,li2016joint} further extended the feature selection method. \citet{ganin2016domain} introduced adversarial learning into domain adaptation. It mapped data from both domains into a domain-invariant domain by fooling a domain discriminator. Some works \cite{he2018adaptive,peng2018cross} introduced semi-supervised learning methods to further utilize the target domain-specific information. Other works \cite{li2017end,li2018hierarchical,hu2019domain,zhang2019interactive} utilized the attention mechanism. Among them, \citet{li2018hierarchical} proposed an effective hierarchical attention transfer network (HATN) that automatically locates pivots and non-pivots in both domains. \citet{zhang2019interactive} put forward an interactive attention transfer network (IATN) to utilize the aspects of documents in cross-domain sentiment classification.\\

\noindent \textbf{Topic model:} Topic model \cite{blei2003latent} is a generative probabilistic model for mining the topic distribution of documents. Recently, variational auto-encoder \cite{DBLP:journals/corr/KingmaW13} based methods have been introduced to topic models which increased the topic construction performance and speed \cite{miao2016neural,miao2017discovering,DBLP:conf/iclr/SrivastavaS17}. Topic model has been applied in many natural language processing (NLP) tasks like text classification \cite{chen2016short,pavlinek2017text,wang2019encoding,wang2020end} while it has not yet been applied in cross-domain sentiment classification. In this paper, we mainly adopt LDA \cite{blei2003latent} as our topic encoding network to assist cross-domain sentiment classification. We have also tried other topic models such as prodLDA \cite{DBLP:conf/iclr/SrivastavaS17} and it turns out that different topic models have little influence on the classification accuracy. Therefore, the experiment results reported in this work are all based on LDA.\\

\noindent \textbf{Attention mechanism in NLP:} The attention mechanism in NLP is firstly used for the machine translation task \cite{DBLP:journals/corr/BahdanauCB14}. It got intuition from the fact that human allocates varied importance to different tokens in a sentence when trying to parse it. Based on the attention mechanism, the transformer \cite{DBLP:conf/nips/VaswaniSPUJGKP17} puts forward a new kind of attention: self-attention, which allows each token to calculate its relative importance with other tokens. The self-attention mechanism has been proven as effective in many tasks, such as language understanding, text classification and semantic role labeling \cite{DBLP:conf/nips/VaswaniSPUJGKP17,DBLP:conf/naacl/DevlinCLT19,tan2018deep}. In this paper, the networks are based on transformers \cite{DBLP:conf/nips/VaswaniSPUJGKP17} and we also utilize the MLP attention mechanism \cite{DBLP:journals/corr/BahdanauCB14} to get fixed-length feature vectors.

\section{Topic Driven Adaptive Network}

\subsection{Problem Definition}
There are two domains in our problem: the source domain $D_s$ and the target domain $D_t$. $X_s$ represents the set of data from $D_s$ and the dataset size is noted as $N_s$. $X_s$ consists of two subsets, i.e., the labeled subset $X_s^l$ and the unlabeled subset $X_s^u$. Each data point in $X_s$ is noted as $x^s_i$ with label $y^s_i$ if having one. Besides, unlabeled dataset $X_t$ from $D_t$ is available and the dataset size is noted as $N_t$. Each data point in $X_t$ is noted as $x^t_i$. A data point in $X_s$ or $X_t$ is noted as $x_i$. In this work, domain-specific words $X^{sp}$ are extracted for each document. For document $x_i$, its corresponding domain-specific words are noted as $x^{sp}_i$.

\subsection{Network Structure}

\begin{figure}[t]
\centering
\includegraphics[width=0.45\textwidth]{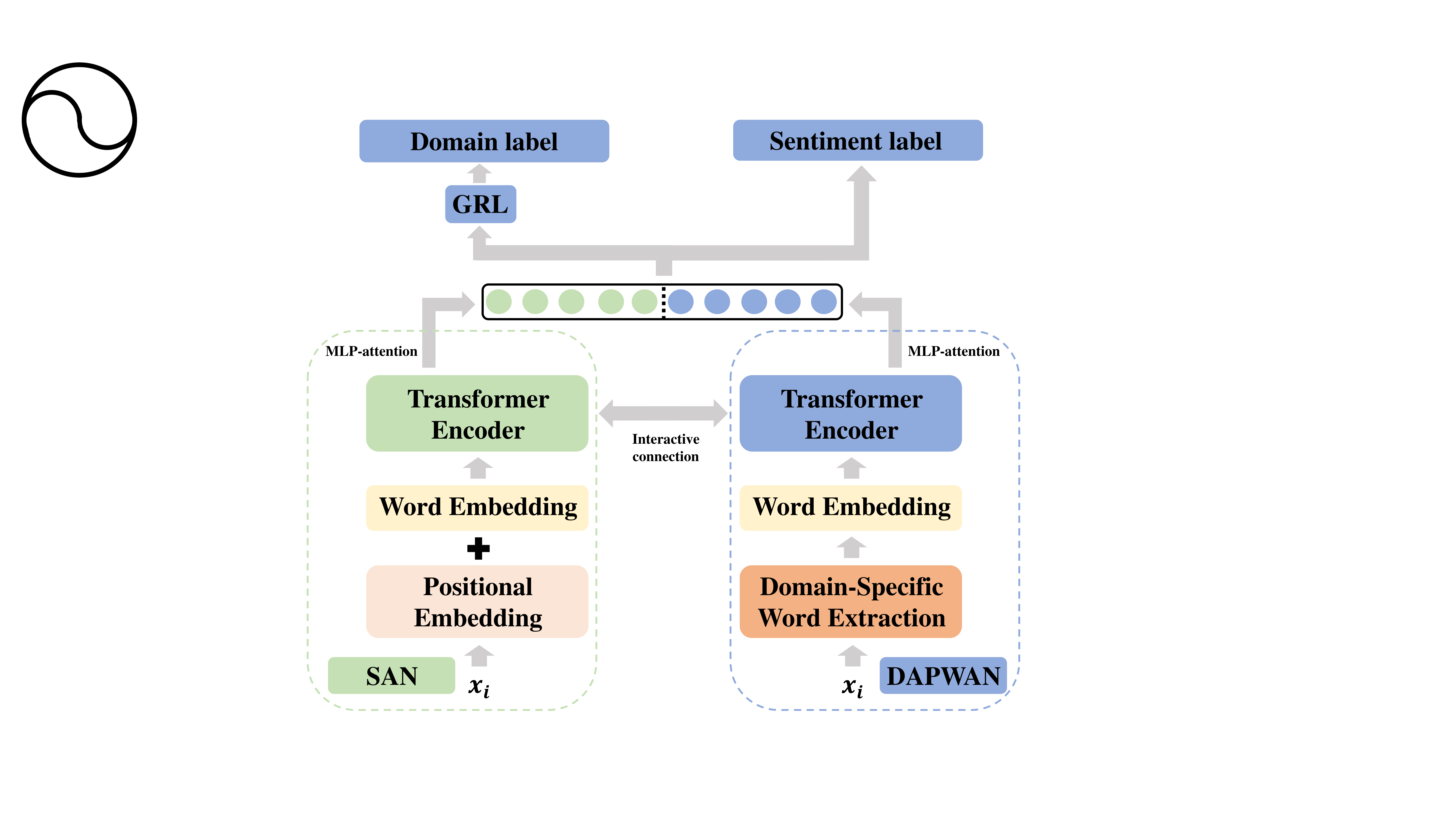}
\caption{The framework of TDAN}
\label{TDAN} 
\end{figure}

Before training the network, we generate topics using documents from both domains. For each cross-domain task, domain-specific words are generated. The domain-specific words are treated as unordered sets. 

As shown in Figure \ref{TDAN}, the proposed TDAN consists of two sub-networks: semantics attention network (SAN) and domain-specific word attention network (DSPWAN). Both SAN and DSPWAN utilize the attention mechanism \cite{DBLP:journals/corr/BahdanauCB14}. Moreover, DSPWAN is interactively connected with SAN to provide more background information as in \citet{zhang2019interactive}.

\subsection{Domain-specific Word Extraction}
\begin{algorithm}[ht]
\caption{Domain-specific word extraction process}
\label{alg:A}
\begin{algorithmic}[1]
\REQUIRE ~~\\ 
The average topic occurrence possibility in source domain, $\mathbf{p}_s$;\\
The average topic occurrence possibility in target domain, $\mathbf{p}_t$;\\
The tolerance bound, $tol$;\\
Documents from source and target domains, $X_s$ and $X_t$;
\ENSURE ~~\\ 
Domain-specific words, $X^{sp}$;

\STATE Initialize two empty topic sets: source domain-specific topic set $s_{sp_s}$ and target domain-specific topic set $s_{sp_t}$;
\FOR{each topic $t$}
\IF{$\mathbf{p}_s[t]-\mathbf{p}_t[t] > tol$}
\STATE Insert $t$ into $s_{sp_s}$;
\ELSIF{$\mathbf{p}_t[t]-\mathbf{p}_s[t] > tol$}
\STATE Insert $t$ into $s_{sp_t}$;
\ENDIF
\ENDFOR
\FOR{each document $x_s^i$ in $X_s$}
    \FOR{each word $w_j$ in $x_s^i$}
        \STATE calculate the most related topic $t_j$ of $w_j$;
        
        \IF{$t_j$ is in $s_{sp_s}$}
        \STATE insert $w_j$ to $x^{sp}_i$;
        \ENDIF
    \ENDFOR
\ENDFOR
\STATE Perform similar operations on $X_t$ to add words to $X^{sp}$;
\end{algorithmic}
\end{algorithm}
In our work, domain-specific words are generated using a topic model. The topic model we adopt here is LDA \cite{blei2003latent}. After the training process, for each document $x_i$, we can obtain its topic distribution vector $p_i \in \mathbb{R}^{k}$, where $k$ is the number of topics. $p_s^i$ represents topic distribution vectors of source document $x_s^i$ and $p_t^i$ represents topic distribution vectors of target document $x_t^i$. The topic model also provides us with the topic-word distribution matrix $\beta\in\mathbb{R}^{V\times k}$, where $V$ is the vocabulary size.

Then, for each cross-domain task, domain-specific words are extracted with the topic model. Algorithm \ref{alg:A} shows the detailed extraction process, where $tol$ is a hyperparameter. For each topic $t$ in the topic model, it is classified as the domain-specific topic if the average possibility difference between two domains is larger than $tol$.  For each word $w_j$ in document $x_i$, it is classified as a domain-specific word if its most related topic is a domain-specific one. The most related topic $t_j$ of a word $w_j$ in document $x_i$ is calculated by:
\begin{equation}
\begin{aligned}
&p_d=\mathbf{\beta}\odot p_i,\\
&t_j = \mathop{\arg\max}_l(p_d[l][w_j]),
\end{aligned}
\end{equation}
where $p_i$ is the topic distribution vector of $x_i$, $\beta$ is the topic-word distribution matrix and $p_d\in\mathbb{R}^{V\times k}$ is the unnormalized word-topic distribution matrix, each row of which denotes the topic distribution of a word in the document. An element-wise multiplication operation is performed between $\beta$ and $p_i$ to generate matrix $p_d$. The most related topic $t_j$ for word $w_j$ is the max element in the corresponding row. 

The average topic occurrence possibilities in two domains, i.e., $\mathbf{p}_s$ and $\mathbf{p}_t$ are calculated as follows:
\begin{equation}
\begin{aligned}
&\mathbf{p}_s=\frac{1}{N_s}\sum_{i=1}^{N_s}p^i_s,\\
&\mathbf{p}_t=\frac{1}{N_t}\sum_{i=1}^{N_t}p^i_t.
\end{aligned}
\end{equation}

After conducting the extraction process in Algorithm \ref{alg:A}, the domain-specific words of a document are obtained. These words are assembled as a sequence and feed into DSPWAN. The domain-specific word sequence corresponding to a document is noted as $x^{sp}_i\in \mathbb{R}^{d_{sp}}$, where $d_{sp}$ is the sequence length. Since the extracted domain-specific words are sparse in a document and their relative order contains little useful information, we remove positional embedding vectors in DSPWAN. 

Also, $x^{sp}_i$ can be empty while empty input for the attention network is not legal. Therefore, a special token \textit{$\langle$specific\_token$\rangle$} is added to $x^{sp}_i$ to smooth the input.

\subsection{Semantics Attention Network}
SAN aims at mining the semantics information from given documents. SAN is based on the attention mechanism. It consists of two components: transformer component \cite{DBLP:conf/nips/VaswaniSPUJGKP17} and MLP attention layer. The transformer is used to extract the context vector $C$ from the input document $x_i$ and the MLP attention layer is used to generate the feature vector used for sentiment classification. For simplicity, we would illustrate the transformer component structure in this section and the MLP-attention layer structure will be introduced in subsequent sections.

In this work, the transformer encoder \cite{DBLP:conf/nips/VaswaniSPUJGKP17} is used as an encoder for documents to generate context vectors.
For each word $w_j$ in the input document $x_i$, it is mapped into the embedding space combining the pre-trained word vector $h_j^{e}$ and the position embedding vector $h_j^{p}$. The embedding vector $h_j\in \mathbb{R}^{d_h}$ of each word is calculated as $h_j=h_j^{e}+h_j^{p}$, where $h_j^{e}\in \mathbb{R}^{d_h}$ and $h_j^{p} \in \mathbb{R}^{d_h}$.  In our experiment, the position embedding is generated as follows:
\begin{equation}
\begin{aligned}
&PE(pos,2l)=sin(pos/10000^{2l/d_{h}}),\\
&PE(pos,2l+1)=cos(pos/10000^{2l/d_{h}}),
\end{aligned}
\end{equation}
where $pos$ is the position and $l$ is the dimension. Each embedding vector $h_j$ of a $d_l$ length document is stacked together to form the matrix $H\in \mathbb{R}^{d_l\times d_h} $. $H$ is encoded using a multi-layer transformer encoder consisting of six multi-head self-attention layers, each followed by a feed-forward layer as in \citet{DBLP:conf/nips/VaswaniSPUJGKP17}. Context vector $C\in \mathbb{R}^{d_l\times d_h} $ is noted as the output of the transformer encoder where each column $c_j$ in $C$ represents the context-aware vector of word $w_j$.


\subsection{Domain-specific Word Attention Network}
DSPWAN aims at encoding domain-specific words to generate the feature vector. Similar to SAN, DSPWAN firstly utilizes the transformer to encode the input into context vectors. Then, an MLP-attention operation is performed on the context vectors to extract the feature vector. Different from SAN, DSPWAN doesn't include positional embedding in the transformer component because the input domain-specific words are unordered. For each word $w^{sp}_j$ in $x^{sp}_{i}$, its embedding vector is noted as $h^{sp}_j\in\mathbb{R}^{d_h}$. Then, the input document $x^{sp}_j$ is mapped into the embedding space as $H^{sp}\in \mathbb{R}^{d_{sp}\times d_h}$, where $H^{sp}$ is stacked using each $h^{sp}_j$.

Then, the context vector $C^{sp}$ is generated from $H^{sp}$ as $C^{sp}=Transformer(H^{sp})$, where $Transformer$ notes the transformer encoder component.

\subsection{Interactive Connection}
SAN and DSPWAN are interactively connected to allow them to incorporate information from each other as in \citet{zhang2019interactive}. Domain-specific words include background information in them and connecting DSPWAN to SAN provides more information for it to utilize. Also, information from SAN enables DSPWAN to better extract useful information from domain-specific words. The specific method is slightly different with \citet{zhang2019interactive} in that the average pooling operation is replaced with an MLP-attention layer. We would introduce the MLP-attention mechanism first before illustrating the interactive connection process.

MLP-attention layer \cite{DBLP:journals/corr/BahdanauCB14} generates a fixed-length feature vector $h \in \mathbb{R}^{d_h}$ using context vector $C$.

Firstly, for every context vector $c_j$ in $C$, the alignment score $f(c_j,q)$ with an abstract query $q\in \mathbb{R}^{d_h} $ is calculated as follows:
\begin{equation}
\begin{aligned}
f(c_j,q)=w^Ttanh(W^{(1)}c_j+W^{(2)}q),
\end{aligned}
\end{equation}
where $q \in\mathbb{R}^{d_h}$,$w\in\mathbb{R}^{d_h\times d_h}$,$W^{(1)}\in\mathbb{R}^{d_h\times d_h}$,$W^{(2)}\in\mathbb{R}^{d_h\times d_h}$ are randomly initialized and updated during the training process. $q$ can be viewed as an high level query on the context vectors about ``which context vector is important".

Secondly, the weight $\alpha_{i}$ of each context vector $c_j$ is computed as follows:
\begin{equation}
\begin{aligned}
\alpha_{i}=\frac{exp(f(c_j,q))}{\sum_{i=1}^{d_l}exp(f(c_j,q))}.
\end{aligned}
\end{equation}

Finally, the feature vector $h$ is calculated as a weighted sum of the context vectors as $h=\sum_{i=1}^{d_l} \alpha_{i} c_j$.

With the MLP-attention layer described above, the pooling vectors of SAN and DSPWAN are generated from context vectors as follows:
\begin{equation}
\begin{aligned}
&h_{sp}^{'}=MLP\_Attention(C),\\
&h_s^{'}=MLP\_Attention(C^{sp}),
\end{aligned}
\end{equation}
where $h_{sp}^{'}$ and $h_s^{'}$ are pooling vectors. The pooling vectors contain high-level information of raw documents and domain-specific words, which are added to context vectors as follows:
\begin{equation}
\begin{aligned}
&C^{'}=concat(C,h_{sp}^{'}),\\
&C_{sp}^{'}=concat(C^{sp},h_s^{'}),
\end{aligned}
\end{equation}
where $concat$ is the function performing concatenation operation. $C^{'}\in\mathbb{R}^{(d_l+1)\times d_h}$ and $C_{sp}^{'}\in\mathbb{R}^{(d_{sp}+1)\times d_h}$ are context vectors containing information from each other. 
The feature vectors of SAN and DSPWAN are generated from $C^{'}$ and $C_{sp}^{'}$ using MLP-attention layers as follows:
\begin{equation}
\begin{aligned}
&h_{sp}=MLP\_Attention(C^{'}),\\
&h_s=MLP\_Attention(C_{sp}^{'}),
\end{aligned}
\end{equation}
where $h_{sp}$ is the output feature vector of DSPWAN and $h_s$ is the output feature vector of SAN.

\subsection{Sentiment Classification}
After generating two feature vectors, we fuse them together and obtain the final feature vector $h_f\in\mathbb{R}^{d_h}$ as follows:
\begin{equation}
\begin{aligned}
h_f=relu(W^{(f)}concat(h_s,h_{sp})+b^{(f)}),
\end{aligned}
\end{equation}
where $W^{(f)}\in\mathbb{R}^{d_h\times (2*d_h)}$ and $b^{(f)}\in\mathbb{R}^{d_h}$ are learnable parameters. Sentiment classification is performed on $h_f$. The sentiment prediction vector $y_c^i\in\mathbb{R}^{2}$ for the $i_{th}$ document is calculated as follows:
\begin{equation}
\begin{aligned}
\hat{y}_c^i=softmax(W^{(c)}h_f+b^{(c)}),
\end{aligned}
\end{equation}
where $W^{(c)}\in\mathbb{R}^{2\times d_h}$ and $b^{(c)}\in\mathbb{R}^{2}$ are learnable parameters. The sentiment classification loss  $L_{cls}$ is defined by:
\begin{equation}
\begin{aligned}
L_{cls}=&-\frac{1}{N_s+N_t}\sum_{i=1}^{N_s+N_t}L_c(y_c^i,\hat{y}_c^i)
\end{aligned}
\end{equation}
where $\hat{y}_c^i\in\mathbb{R}^{2}$ is the ground truth label and $L_c$ is the cross entropy loss function. The sentiment classifier is trained by minimizing $L_{cls}$. 

\subsection{Domain Classification}
The domain classifier in TDAN aims to distinguish feature vectors $h_f$ from different domains. The domain prediction vector $y_d^i\in\mathbb{R}^{2}$ for the $i_{th}$ document is calculated as follows:
\begin{equation}
\begin{aligned}
y_d=softmax(W^{(d)}h_f+b^{(d)}),
\end{aligned}
\end{equation}
where $W^{(d)}\in\mathbb{R}^{2\times d_h}$ and $b^{(d)}\in\mathbb{R}^{2}$ are learnable parameters of the domain classifier. Domain classification loss $L_{dom}$ is defined by the cross-entropy loss:
\begin{equation}
\begin{aligned}
L_{dom}=&-\frac{1}{N_s+N_t}\sum_{i=1}^{N_s+N_t}L_c(y_d^i,\hat{y}_d^i),
\end{aligned}
\end{equation}
where $\hat{y}_d^i\in\{0,1\}$. $\hat{y}_d^i$ is the ground truth label and $L_c$ is the cross entropy loss function. The domain classifier is trained by minimizing $L_{dom}$. 

Besides, a Gradient Reversal Layer (GRL) is added before the domain classifier \cite{ganin2016domain}.  Mathematically, GRL is defined as $Q_\lambda(x)=x$ with a reversal gradient $\frac{\partial Q_\lambda(x)}{\partial x}=-\lambda x$. Training with GRL is adversarial because the domain classifier would try to distinguish feature vectors from different domains while the reversed gradient helps the generator produce indistinguishable feature vectors. This is how domain adaptation is performed in our work. Ideally, the generator in TDAN would generate feature vectors in a common feature space.

\subsection{Training Strategy}
Our model is trained by minimizing the loss item $L_{total}$, which is calculated as $L_{total}=L_{cls}+\rho L_{dom}$. $\rho$ is a hyperparameter to balance the relative importance of loss items. The sentiment classification is performed only in source labeled data, therefore $L_{cls}$ is calculated with $X_s^l$. The domain classification utilizes all data points in $X_s^l$ and $X_t$, therefore $L_{dom}$ is calculated with $X_s^l$ and $X_t$. Each mini-batch in training steps contains balanced data points from the source and target domains. 

\section{Experiments}

\subsection{Datasets}
\begin{table*}[t]
\centering

\begin{tabular}{c|ccccccc|ccc}
\hline
        &  \multicolumn{7}{c|}{Baselines} &        \multicolumn{3}{c}{Our Methods}  \\ \hline
  Task  &  SCL & SFA & DANN     & HATN$_h$ & IATN &ACAN &TPT            &TDAN-& TDAN$_f$ & TDAN\\ \hline \hline
B$\to$ D &0.803 & 0.826 & 0.827 & 0.863 & 0.868 &0.865 &0.828       & \textbf{0.871} & 0.868 & 0.869 \\
B$\to$E &0.742 & 0.771 & 0.804  & 0.851 & 0.854   &0.8278&0.806     & 0.848 & 0.853 & \textbf{0.859}  \\
B$\to$K &0.774 & 0.786 & 0.843  & 0.852 & 0.859   & 0.848 &0.833    & 0.842  & 0.861 & \textbf{0.862} \\ \hline
D$\to$B &0.764 & 0.802 & 0.825  & 0.865 & 0.859  &0.8622 &0.826     & \textbf{0.868} & 0.864 & 0.864\\ 
D$\to$E &0.740 & 0.765 & 0.809  & 0.856 & 0.842  &0.8342 &0.819     & 0.844 & 0.852 & \textbf{0.854}\\
D$\to$K &0.757 & 0.778& 0.849   & 0.865 & 0.858&0.840  &0.824       & 0.841 & 0.858 & \textbf{0.866} \\ \hline
E$\to$B &0.703 & 0.727& 0.774   & 0.819 & 0.809 &  0.8184 &0.772    & 0.821 & 0.829 & \textbf{0.840} \\
E$\to$D &0.724 & 0.756& 0.781   & 0.838 & 0.815 &  0.825 &0.812     & 0.816 & 0.839 & \textbf{0.842}\\
E$\to$K &0.823 & 0.845& 0.881   & 0.886 & 0.891 &  0.899 &0.867     & 0.895 & 0.890 & \textbf{0.897}\\ \hline
K$\to$B &0.704 & 0.733& 0.718   & 0.835 & 0.836 & 0.8182 &0.771     & 0.832 & 0.842 & \textbf{0.844}\\
K$\to$D &0.737 & 0.758& 0.789   & 0.841 & 0.839 & 0.8274 &0.799     & 0.823 & 0.836 & \textbf{0.844}\\
K$\to$E &0.824 & 0.844& 0.846   & 0.878 & 0.882 &  0.8784 &0.874    & 0.877 & 0.878 & \textbf{0.893}\\ \hline \hline
AVG     &0.758 & 0.782& 0.812   & 0.854 & 0.850 &  0.845 &0.819     & 0.849 & 0.856 & \textbf{0.861}\\ \hline
\end{tabular}
\caption{Classification accuracy on the Amazon reviews dataset.}
\label{table:clsAcu}
\end{table*}

\begin{table*}[t]
\centering

\begin{tabular}{c|ccccccc|ccc}
\hline
        &  \multicolumn{7}{c|}{Baselines} &        \multicolumn{3}{c}{Our Methods}\\ \hline
  Task  &  SCL & SFA & DANN         & HATN$_h$ & IATN &ACAN &TPT      &TDAN-& TDAN$_f$ & TDAN\\ \hline \hline
B$\to$Y &0.776 & 0.768 & 0.835    & 0.868 & 0.869 & 0.852 & 0.850     & 0.862 & 0.870 & \textbf{0.871} \\
D$\to$Y &0.749 & 0.791 & 0.815    & 0.869 & 0.871 & 0.876 & 0.870     & 0.872 & \textbf{0.882} & 0.876\\ 
E$\to$Y &0.747 & 0.743 & 0.803    & 0.873 & 0.852 & 0.872 & 0.864     & 0.874 & \textbf{0.889} & 0.885 \\
K$\to$Y &0.753 & 0.793 & 0.821    & 0.869 & 0.884 & 0.872 & 0.840     & 0.880 & 0.883 & \textbf{0.887}\\\hline
Y$\to$B &0.712 & 0.735 & 0.750    & 0.821 & 0.819 & 0.809 & 0.762     & 0.809 & 0.818 & \textbf{0.823}\\
Y$\to$D &0.704 & 0.704 & 0.756    & 0.812 & 0.818 & 0.813 & 0.752     & 0.825 & 0.825 & \textbf{0.833} \\
Y$\to$E &0.738 & 0.761 & 0.775    & 0.841 & \textbf{0.859} & 0.823 & 0.751     & 0.833 & 0.840 & 0.844 \\ 
Y$\to$K &0.748 & 0.791 & 0.783    & 0.851 & 0.857 & 0.834 & 0.796     & 0.842 & 0.851 & \textbf{0.863}\\ \hline \hline
AVG     &0.741 & 0.761 & 0.792    & 0.851 & 0.854 & 0.843 & 0.810     & 0.850 & 0.857 & \textbf{0.860}\\ \hline
\end{tabular}
\caption{Classification accuracy between Amazon reviews dataset and Yelp reviews dataset.}
\label{table:clsYA}
\end{table*}

Our network is evaluated on the Amazon reviews dataset \cite{blitzer7domain} and the Yelp reviews dataset\footnote{https://www.yelp.com/dataset}. The Amazon reviews dataset contains reviews from four domains: Book(B), DVD (D), Electronics (E), and Kitchen appliances (K). Each domain contains 6000 reviews, where 3000 reviews are positive (higher than 3 stars) and 3000 reviews are negative (lower than 3 stars). Besides, each domain contains lots of unlabeled data. The Yelp reviews dataset contains 8635403 restaurant reviews and each review is attached with a score from 1 to 5. A review is treated as positive if its score is higher than 3 and negative if lower or equal to 3. The Yelp reviews dataset is treated as a unique domain (Y). In experiments, 3000 positive reviews and 3000 negative reviews are randomly selected from the Yelp dataset to keep the same domain data size as the Amazon reviews dataset. Based on the five domains in two datasets, we construct twenty cross-domain tasks, each noted by A $\to$ B where A is the source domain and B is the target domain. The cross-domain tasks between two datasets can help avoid bias on a single dataset.



\subsection{Experimental Settings}
In our experiment, word embedding dimension $d_w$ is set to 300 and we adopt pretrained word embeddings provided by Google\footnote{https://code.google.com/archive/p/word2vec/}. The dimension of hidden vector $d_h$ is set to 300. The tolerance bound $tol$ is set to 0.08. The topic number $k$ is set to 50. The hyperparameter $\rho$ is set to 1. All model weights are xavier initialized \cite{glorot2010understanding}. Model weights are optimized using the ADAM optimizer \cite{DBLP:journals/corr/KingmaB14} with a learning rate of 0.0002 and weight decay rate of 5e-5. Dropout \cite{srivastava2014dropout} is adopted and the dropout rate is set to 0.25. The adaptation rate $\lambda$ is increased as $\lambda=min(\frac{2}{1+exp(-10\frac{t}{T})}-1,0.1)$ where $t$ is the current epoch, and $T$ is the maximum epoch which is set to 50. The self-attention layer number is 6 for SAN and 3 for DSPWAN. The multi-head number of self-attention is 4 for each sub-network.

The model is trained using mini-batch. Considering the varied size of data points needed for calculating each loss, a batch size of 40 is used, with 20 data points coming from $X_s$ and 20 data points coming from $X_t$. $L_{cls}$ is calculated on the 20 data points from $X_s$. $L_{dom}$ are calculated on the full mini-batch. 

We randomly select 1000 data points from the target domain as the development set and the other 5000 data points as the testing set. The positive and negative instances in the development and testing sets are balanced. We perform early stop in training when evaluation results no longer increase for 10 epochs. Also, the hyperparameters mentioned above are selected using the development set.

\subsection{Baselines}
\begin{table*}[t]

\centering
\begin{tabular}{l|l}
\hline
\textbf{books domain:} & \makecell[l]{\textbf{kitchen domain:}}\\ \hline\hline
\makecell[l]{\textbf{document:}\\
\colorbox[rgb]{0.843 , 0.914 , 0.996}{should} 
be 
\colorbox[rgb]{0.843 , 0.914 , 0.996}{recommended}
reading. for everyone\\ 
\colorbox[rgb]{0.639 , 0.812 , 0.980}{interested}
in this subject. patrick heron has done\\
an \colorbox[rgb]{0.482 , 0.729 , 0.976}{excellent}
job explaining the nephilim.\\ this book has answered 
\colorbox[rgb]{0.843 , 0.914 , 0.996}{all}
my 
\colorbox[rgb]{0.639 , 0.812 , 0.980}{questions}
.}
&
\makecell[l]{\textbf{document:}\\
\colorbox[rgb]{0.482 , 0.729 , 0.976}{best}
thing ever. one cup of coffee at a 
\colorbox[rgb]{0.639 , 0.812 , 0.980}{time}
\colorbox[rgb]{0.639 , 0.812 , 0.980}{ doesn't}\\
get 
\colorbox[rgb]{0.843 , 0.914 , 0.996}{better}
than this. 
\colorbox[rgb]{0.482 , 0.729 , 0.976}{perfect}
for a household\\ of coffee drinkers that wake  up at different times} 
 \\  \hline
\makecell[l]{\textbf{domain-specific words:}\\ 
\colorbox[rgb]{0.843 , 0.914 , 0.996}{reading}
\colorbox[rgb]{0.482 , 0.729 , 0.976}{interested}
patrick heron nephilim\\ 
\colorbox[rgb]{0.843 , 0.914 , 0.996}{book}
subject 
\colorbox[rgb]{0.843 , 0.914 , 0.996}{questions}
} & \makecell[l]{\textbf{domain-specific words:}\\ cup 
\colorbox[rgb]{0.482 , 0.729 , 0.976}{coffee}
\colorbox[rgb]{0.639 , 0.812 , 0.980}{drinkers}
}  \\ \hline
\end{tabular}
\caption{Attention visualization of two samples in B$\to$K task. The left column is a document from the books domain and the right column is a document from the kitchen domain. Their domain-specific words are also included.}
\label{table:AV}
\end{table*}
\begin{itemize}

\item\textbf{SCL} \cite{blitzer7domain} induces correspondences among features
from different domains for sentiment classfication.
\item\textbf{SFA} \cite{pan2010cross} is a linear method which aligns no-pivots from different domains into unified clusters via pivots.
\item\textbf{DANN} \cite{ganin2016domain} is an adversarial domain-adaptation method. It performs domain adaptation on reviews encoded in 5000 dimensional feature vectors of unigrams and bigrams.
\item\textbf{HATN$^h$} \cite{li2018hierarchical} is a hierarchy attention network for cross-domain sentiment classification. Considering that HATN$^h$ performs better than HATN in \citet{li2018hierarchical}, it is used in our experiments.
\item\textbf{IATN} \cite{zhang2019interactive} is an attention-based network which combines aspect information to help with the sentiment classification task. 
\item\textbf{ACAN} \cite{qu2019adversarial} is a method which enhances category consistency between the source domain and the target domain using two seperate classifiers.
\item\textbf{TPT} \cite{li2020simultaneous} is a method of learning pivots and representations simultaneously using a pivot selector and a transferable transformer. Sentiment classification is performed on these domain-invariant representations.
\end{itemize}
\subsection{Result Analysis}

Following the typical experiment settings in previous works \cite{li2017end,li2018hierarchical,zhang2019interactive}, we adopt classification accuracy as the evaluation metric. Table \ref{table:clsAcu} shows cross-domain task results on the Amazon review dataset and Table \ref{table:clsYA} shows the cross-domain task results between two datasets. TDAN$_f$ is a variant of TDAN by simply fusing $h_s$ with $h_{sp}$ without applying the interactive interaction. TDAN- is a variant of TDAN by only using SAN. According to Table \ref{table:clsAcu} and Table \ref{table:clsYA}, TDAN outperforms previous methods. The traditional method SCL and SFA show limited performance. The methods based on adversarial domain adaptation and deep learning witness significant performance improvement, most of which obtain over 80\% average classification accuracy.
The model proposed by us shows a significant increase in performance. TDAN outperforms DANN by 4.9\%, HATN$_h$ by 0.7\%, IATN by 1.1\%, ACAN by 1.6\%, TPT by 4.2\% on the Amazon reviews dataset. It outperforms DANN by 6.8\%, HATN$_h$ by 0.9\%, IATN by 0.6\%, ACAN by 1.7\%, TPT by 5\% on the cross-dataset experiments. 

We have also compared our model with the method based on pre-training language models. BERT-DAAT \cite{du2020adversarial} is a cross-domain sentiment classification method based on BERT and we evaluate its performance on the Amazon reviews dataset and the Yelp reviews dataset. It achieves the average classification accuracy of 89.7\% on the Amazon reviews dataset and 89.4\% on the Yelp reviews dataset. However, it requires much more training time and graphic card memory. A single cross-domain task in BERT-DAAT costs about 3 hours to finish on two Nvidia V100 graphic cards (32GB memory for each card) while TDAN costs 1 hour for a single cross-domain task using two Nvidia 1080Ti graphic cards (11GB memory for each card). Therefore, our model saves training time and device memory.

To validate the effect of our improvement on TDAN-, we compare several variants of TDAN with it. First of all, TDAN$_f$ outperforms TDAN- by 0.7\% on both datasets. The reason behind this is that DSPWAN can utilize both kinds of domain-specific words, i.e., non-pivots and background words, to assist with sentiment classification. Moreover, TDAN outperforms TDAN$_f$ by 0.5\% and 0.3\%, showing that the introduction of interactive connection enables sub-networks to gain information from each other and thus improve the model performance.

\subsection{Attention Visualization}
To show that TDAN can develop different attention strategies for different forms of input, we visualize the attention score of the MLP-attention layer in Table \ref{table:AV}. We can see that the semantics attention network pays attention to pivots, such as \textit{excellent} and \textit{perfect}. It also notices words that affect the sentence semantics, such as \textit{doesn't}. The domain-specific word attention network pays attention to both background words and non-pivots. For example, the right document in Table \ref{table:AV} does not contain non-pivots in its domain-specific words and the domain-specific word attention network pays attention to background words such as \textit{coffee} and \textit{drinkers}. However, the left document in Table \ref{table:AV} contains non-pivots \textit{interested} therefore the domain-specific word attention network pays most attention to it while allocating lower attention scores to other background words. This ability to pay attention to both background words and non-pivots is crucial for sentiment classification.
\begin{table}[t]

\begin{tabular}{l|l|l|l}
\hline
& HATN      & IATN & TDAN \\ \hline \hline
Amazon& 0.859 &    0.855       &0.861              \\\hline
Amazon and Yelp & 0.857       & 0.856           & 0.859       \\\hline

\end{tabular}
\caption{Influence of different domain-specific word extraction methods on classification accuracy. The average accuracy on two datasets are reported.}
\label{table:SP}
\end{table}
\subsection{Influnce of Domain-specific Word Extraction}
To show the effect of our topic-based domain-specific word extraction method, we compare it with the other two methods from HATN and IATN. In HATN, non-pivots in NP-net are treated as domain-specific words. In IATN, aspects are treated as domain-specific words. To be fair, we use the same model, TDAN, to encode the input generated using different methods. The average accuracy results are reported in Table \ref{table:SP}. It has been shown that our methods witness improvements in classification accuracy. The reason is that our method includes both non-pivots and background words in the extracted domain-specific words for sentiment classification. 
\section{Conclusions}
In this work, we proposed a novel network TDAN for cross-domain sentiment classification. A new domain-specific word extraction method based on the topic information was proposed, aiming to assist sentiment classification. Specifically, both types of domain-specific words are included. Additionally, we utilize both self-attention and MLP-attention mechanisms to build TDAN. The experiment results witness improved classification performance on two datasets. Also, the experiment results show that our domain-specific word extraction method outperforms previous ones.
\bibliography{aaai22}

\begin{thebibliography}{33}
\providecommand{\natexlab}[1]{#1}

\bibitem[{Bahdanau, Cho, and Bengio(2015)}]{DBLP:journals/corr/BahdanauCB14}
Bahdanau, D.; Cho, K.; and Bengio, Y. 2015.
\newblock Neural machine translation by jointly learning to align and
  translate.
\newblock In \emph{Proceedings of the 3rd International Conference on Learning
  Representations}.

\bibitem[{Blei, Ng, and Jordan(2003)}]{blei2003latent}
Blei, D.~M.; Ng, A.~Y.; and Jordan, M.~I. 2003.
\newblock Latent dirichlet allocation.
\newblock \emph{Journal of Machine Learning Research}, 3: 993--1022.

\bibitem[{Blitzer, Dredze, and Pereira(2007)}]{blitzer7domain}
Blitzer, J.; Dredze, M.; and Pereira, F. 2007.
\newblock Biographies, bollywood, boom-boxes and blenders: Domain adaptation
  for sentiment classification.
\newblock In \emph{Proceedings of the 45th Annual Meeting of the Association
  for Computational Linguistics}, 440--447.

\bibitem[{Blitzer, McDonald, and Pereira(2006)}]{blitzer-etal-2006-domain}
Blitzer, J.; McDonald, R.~T.; and Pereira, F. 2006.
\newblock Domain adaptation with structural correspondence learning.
\newblock In \emph{Proceedings of the 2006 Conference on Empirical Methods in
  Natural Language Processing}, 120--128.

\bibitem[{Devlin et~al.(2019)Devlin, Chang, Lee, and
  Toutanova}]{DBLP:conf/naacl/DevlinCLT19}
Devlin, J.; Chang, M.; Lee, K.; and Toutanova, K. 2019.
\newblock {BERT:} Pre-training of deep bidirectional transformers for language
  understanding.
\newblock In \emph{Proceedings of the 2019 Conference of the North American
  Chapter of the Association for Computational Linguistics: Human Language
  Technologies}, 4171--4186.

\bibitem[{Du et~al.(2020)Du, Sun, Wang, Qi, and Liao}]{du2020adversarial}
Du, C.; Sun, H.; Wang, J.; Qi, Q.; and Liao, J. 2020.
\newblock Adversarial and domain-aware {BERT} for cross-domain sentiment
  analysis.
\newblock In \emph{Proceedings of the 58th Annual Meeting of the Association
  for Computational Linguistics}, 4019--4028.

\bibitem[{Ganin et~al.(2016)Ganin, Ustinova, Ajakan, Germain, Larochelle,
  Laviolette, Marchand, and Lempitsky}]{ganin2016domain}
Ganin, Y.; Ustinova, E.; Ajakan, H.; Germain, P.; Larochelle, H.; Laviolette,
  F.; Marchand, M.; and Lempitsky, V.~S. 2016.
\newblock Domain-adversarial training of neural networks.
\newblock \emph{Journal of Machine Learning Research}, 17: 59:1--59:35.

\bibitem[{Glorot and Bengio(2010)}]{glorot2010understanding}
Glorot, X.; and Bengio, Y. 2010.
\newblock Understanding the difficulty of training deep feedforward neural
  networks.
\newblock In \emph{Proceedings of the 13th International Conference on
  Artificial Intelligence and Statistics}, 249--256.

\bibitem[{Goodfellow et~al.(2014)Goodfellow, Pouget{-}Abadie, Mirza, Xu,
  Warde{-}Farley, Ozair, Courville, and Bengio}]{NIPS2014_5ca3e9b1}
Goodfellow, I.~J.; Pouget{-}Abadie, J.; Mirza, M.; Xu, B.; Warde{-}Farley, D.;
  Ozair, S.; Courville, A.~C.; and Bengio, Y. 2014.
\newblock Generative adversarial nets.
\newblock In \emph{Proceedings of the 27th Annual Conference on Neural
  Information Processing Systems}, 2672--2680.

\bibitem[{Gu, Li, and Han(2011)}]{gu2011joint}
Gu, Q.; Li, Z.; and Han, J. 2011.
\newblock Joint feature selection and subspace learning.
\newblock In \emph{Proceedings of the 22nd International Joint Conference on
  Artificial Intelligence}, 1294--1299.

\bibitem[{He et~al.(2018)He, Lee, Ng, and Dahlmeier}]{he2018adaptive}
He, R.; Lee, W.~S.; Ng, H.~T.; and Dahlmeier, D. 2018.
\newblock Adaptive semi-supervised learning for cross-domain sentiment
  classification.
\newblock In \emph{Proceedings of the 2018 Conference on Empirical Methods in
  Natural Language Processing}, 3467--3476.

\bibitem[{Hu et~al.(2019)Hu, Wu, Zhao, Guo, Cheng, and Su}]{hu2019domain}
Hu, M.; Wu, Y.; Zhao, S.; Guo, H.; Cheng, R.; and Su, Z. 2019.
\newblock Domain-invariant feature distillation for cross-Domain sentiment
  classification.
\newblock In \emph{Proceedings of the 2019 Conference on Empirical Methods in
  Natural Language Processing and the 9th International Joint Conference on
  Natural Language Processing}, 5558--5567.

\bibitem[{Kingma and Ba(2015)}]{DBLP:journals/corr/KingmaB14}
Kingma, D.~P.; and Ba, J. 2015.
\newblock Adam: {A} method for stochastic optimization.
\newblock In \emph{Proceedings of the 3rd International Conference on Learning
  Representations}.

\bibitem[{Kingma and Welling(2014)}]{DBLP:journals/corr/KingmaW13}
Kingma, D.~P.; and Welling, M. 2014.
\newblock Auto-encoding variational bayes.
\newblock In \emph{Proceedings of the 2nd International Conference on Learning
  Representations}.

\bibitem[{Li, Zhao, and Lu(2016)}]{li2016joint}
Li, J.; Zhao, J.; and Lu, K. 2016.
\newblock Joint feature selection and structure preservation for domain
  adaptation.
\newblock In \emph{Proceedings of the 25th International Joint Conference on
  Artificial Intelligence}, 1697--1703.

\bibitem[{Li et~al.(2020)Li, Ye, Long, Tang, Xu, and Wang}]{li2020simultaneous}
Li, L.; Ye, W.; Long, M.; Tang, Y.; Xu, J.; and Wang, J. 2020.
\newblock Simultaneous learning of pivots and representations for cross-domain
  sentiment classification.
\newblock In \emph{Proceedings of the Thirty-Fourth {AAAI} Conference on
  Artificial Intelligence}, 8220--8227.

\bibitem[{Li et~al.(2018)Li, Wei, Zhang, and Yang}]{li2018hierarchical}
Li, Z.; Wei, Y.; Zhang, Y.; and Yang, Q. 2018.
\newblock Hierarchical attention transfer network for cross-domain sentiment
  classification.
\newblock In \emph{Proceedings of the 32nd {AAAI} Conference on Artificial
  Intelligence}, 5852--5859.

\bibitem[{Li et~al.(2017)Li, Zhang, Wei, Wu, and Yang}]{li2017end}
Li, Z.; Zhang, Y.; Wei, Y.; Wu, Y.; and Yang, Q. 2017.
\newblock End-to-End Adversarial Memory Network for Cross-domain Sentiment
  Classification.
\newblock In Sierra, C., ed., \emph{Proceedings of the 26th International Joint
  Conference on Artificial Intelligence}, 2237--2243.

\bibitem[{Miao, Grefenstette, and Blunsom(2017)}]{miao2017discovering}
Miao, Y.; Grefenstette, E.; and Blunsom, P. 2017.
\newblock Discovering discrete latent topics with neural variational inference.
\newblock In \emph{Proceedings of the 34th International Conference on Machine
  Learning}, 2410--2419.

\bibitem[{Miao, Yu, and Blunsom(2016)}]{miao2016neural}
Miao, Y.; Yu, L.; and Blunsom, P. 2016.
\newblock Neural variational inference for text processing.
\newblock In \emph{Proceedings of the 33nd International Conference on Machine
  Learning}, 1727--1736.

\bibitem[{Pan et~al.(2010)Pan, Ni, Sun, Yang, and Chen}]{pan2010cross}
Pan, S.~J.; Ni, X.; Sun, J.; Yang, Q.; and Chen, Z. 2010.
\newblock Cross-domain sentiment classification via spectral feature alignment.
\newblock In \emph{Proceedings of the 19th International Conference on World
  Wide Web}, 751--760.

\bibitem[{Pavlinek and Podgorelec(2017)}]{pavlinek2017text}
Pavlinek, M.; and Podgorelec, V. 2017.
\newblock Text classification method based on self-training and {LDA} topic
  models.
\newblock \emph{Expert Systems with Applications}, 80: 83--93.

\bibitem[{Peng et~al.(2018)Peng, Zhang, Jiang, and Huang}]{peng2018cross}
Peng, M.; Zhang, Q.; Jiang, Y.; and Huang, X. 2018.
\newblock Cross-domain sentiment classification with target domain specific
  information.
\newblock In \emph{Proceedings of the 56th Annual Meeting of the Association
  for Computational Linguistics}, 2505--2513.

\bibitem[{Qiuxing, Lixiu, and Jie(2016)}]{chen2016short}
Qiuxing, C.; Lixiu, Y.; and Jie, Y. 2016.
\newblock Short text classification based on LDA topic model.
\newblock In \emph{Proceedings of the 2016 International Conference on Audio,
  Language and Image Processing}, 749--753.

\bibitem[{Qu et~al.(2019)Qu, Zou, Cheng, Yang, and Zhou}]{qu2019adversarial}
Qu, X.; Zou, Z.; Cheng, Y.; Yang, Y.; and Zhou, P. 2019.
\newblock Adversarial category alignment network for cross-domain sentiment
  classification.
\newblock In \emph{Proceedings of the 2019 Conference of the North American
  Chapter of the Association for Computational Linguistics: Human Language
  Technologies}, 2496--2508.

\bibitem[{Srivastava and Sutton(2017)}]{DBLP:conf/iclr/SrivastavaS17}
Srivastava, A.; and Sutton, C. 2017.
\newblock Autoencoding variational inference for topic models.
\newblock In \emph{Proceedings of the 5th International Conference on Learning
  Representations}.

\bibitem[{Srivastava et~al.(2014)Srivastava, Hinton, Krizhevsky, Sutskever, and
  Salakhutdinov}]{srivastava2014dropout}
Srivastava, N.; Hinton, G.~E.; Krizhevsky, A.; Sutskever, I.; and
  Salakhutdinov, R. 2014.
\newblock Dropout: a simple way to prevent neural networks from overfitting.
\newblock \emph{Journal of Machine Learning Research}, 15(1): 1929--1958.

\bibitem[{Tan et~al.(2018)Tan, Wang, Xie, Chen, and Shi}]{tan2018deep}
Tan, Z.; Wang, M.; Xie, J.; Chen, Y.; and Shi, X. 2018.
\newblock Deep semantic role labeling With self-attention.
\newblock In \emph{Proceedings of the Thirty-Second {AAAI} Conference on
  Artificial Intelligence}, 4929--4936.

\bibitem[{Vaswani et~al.(2017)Vaswani, Shazeer, Parmar, Uszkoreit, Jones,
  Gomez, Kaiser, and Polosukhin}]{DBLP:conf/nips/VaswaniSPUJGKP17}
Vaswani, A.; Shazeer, N.; Parmar, N.; Uszkoreit, J.; Jones, L.; Gomez, A.~N.;
  Kaiser, L.; and Polosukhin, I. 2017.
\newblock Attention is all you need.
\newblock In \emph{Proceedings of the 30th Annual Conference on Neural
  Information Processing Systems}, 5998--6008.

\bibitem[{Wang and Wang(2020)}]{wang2020end}
Wang, C.; and Wang, B. 2020.
\newblock An end-to-end topic-enhanced self-Attention network for social
  emotion classification.
\newblock In \emph{Proceedings of the Web Conference 2020}, 2210--2219.

\bibitem[{Wang et~al.(2019)Wang, Wang, Xiang, and Xu}]{wang2019encoding}
Wang, C.; Wang, B.; Xiang, W.; and Xu, M. 2019.
\newblock Encoding syntactic dependency and topical information for social
  emotion classification.
\newblock In \emph{Proceedings of the 42nd International {ACM} {SIGIR}
  Conference on Research and Development in Information Retrieval}, 881--884.

\bibitem[{Zhang et~al.(2019)Zhang, Zhang, Liu, Zhao, Zhu, and
  Chen}]{zhang2019interactive}
Zhang, K.; Zhang, H.; Liu, Q.; Zhao, H.; Zhu, H.; and Chen, E. 2019.
\newblock Interactive attention transfer betwork for cross-Domain sentiment
  classification.
\newblock In \emph{The Thirty-Third {AAAI} Conference on Artificial
  Intelligence}, 5773--5780.

\bibitem[{Zhang, Wang, and Liu(2018)}]{DBLP:journals/widm/ZhangWL18}
Zhang, L.; Wang, S.; and Liu, B. 2018.
\newblock Deep learning for sentiment analysis: {A} survey.
\newblock \emph{Wiley Interdisciplinary Reviews: Data Mining and Knowledge
  Discovery}, 8(4).

\end{thebibliography}
\end{document}